\documentclass[conference]{IEEEtran}
\usepackage{cite}

\ifCLASSINFOpdf
  \usepackage[pdftex]{graphicx}
\else
\fi

\usepackage{amsmath}
\usepackage{amssymb}
\usepackage{diagbox}
\usepackage{subfig}
\usepackage{tikz}
\usetikzlibrary{shadows.blur}
\usepackage{booktabs}

\newcommand{\latin}[1]{{\it #1}}

\usepackage{url}
\usepackage{breakurl}
\usepackage[breaklinks]{hyperref}

\newcounter{nbdrafts}
\makeatletter
\newcommand{\checknbdrafts}{
\ifnum \thenbdrafts > 0
\@latex@warning@no@line{**********************************************************************}
\@latex@warning@no@line{* The document contains \thenbdrafts \space draft note(s)}
\@latex@warning@no@line{**********************************************************************}
\fi}

\makeatother

\begin{document}
\title{Deep Multi-Camera People Detection}

\author{\IEEEauthorblockN{Tatjana Chavdarova and
Fran\c cois Fleuret}
\IEEEauthorblockA{
Idiap Research Institute and \\  \'Ecole Polytechnique F\'ed\'erale de Lausanne}
Email: \{firstname.lastname\}@idiap.ch
}
\maketitle

\begin{abstract}

This paper addresses the problem of multi-view people occupancy map
estimation.  Existing solutions for this problem either operate
per-view, or rely on a background subtraction pre-processing. Both
approaches lessen the detection performance as scenes become more
crowded. The former does not exploit joint information, whereas
the latter deals with ambiguous input due to the foreground blobs
becoming more and more interconnected as the number of targets
increases.

Although deep learning algorithms have proven to excel on remarkably
numerous computer vision tasks, such a method has not been applied yet to this
problem.  In large part this is due to the lack of large-scale
multi-camera data-set.

The core of our method is an architecture which makes use of monocular
pedestrian data-set, available at larger scale then the multi-view
ones, applies parallel processing to the multiple video streams, and
jointly utilises it.  Our end-to-end deep learning method outperforms
existing methods by large margins on the commonly used PETS 2009
data-set.  Furthermore, we make publicly available a new three-camera HD
data-set. Our source code and trained models will be made available 
under an open-source license.

\end{abstract}
\IEEEpeerreviewmaketitle

\section{Introduction}

Due to the high demand in applications, pedestrian detection differentiated itself as a separate class from object detection, enjoying separate attention from the research community.
In spite of recent advances, performance of monocular methods remains limited due to the occlusions that often occur among the individuals. Multi-camera approaches offer a promising extension to resolve the one-view detection ambiguities, provided that the multi-stream information is used jointly to yield the detection estimation.
In this paper, we confine our discussion to a set-up of either a single calibrated static
camera, or several synchronized cameras with overlapping fields of view.

Surprisingly, currently only few multi-view methods jointly utilize
the information across views, which for short we refer to as ``joint methods.'' Every existing joint method performs background subtraction pre-processing of the input images, where the goal is to segment the moving objects out of the background, while taking into account pixel-wise time consistency. There exist a vast catalog of background-subtraction methods, specific to certain applications or even illumination regimes, often difficult to tune, error-prone and noisy. For the multi-camera people detection problem in particular, the background subtraction methods introduce ambiguities when the foreground segmented blobs are interconnected, which limits the success of the joint methods to less-crowded applications.
In addition, there may be moving objects in the scene which are not person and are still segmented.
On the other hand, the state of the art monocular people detection methods leverage the full signal either by using a deep Convolutional Neural Network (CNN), or by building predictors on top of ``deep features'' extracted by such a network.
Surprisingly, the multi-view occupancy map estimation problem has not been re-visited to incorporate such techniques.

On the contrary, the ongoing research on multi-camera people detection incorporates hand-crafted features,
whereas to the best of our knowledge real-world industrial applications use monocular CNN pedestrian detector, and the final estimation is done by averaging the separate per-view ones.
We presume that this is mostly due to the non-existence of  large-scale multi-camera data-set which would allow for training a multi-input architecture.

The method we propose in this paper consists of:
(1) fine-tuning a state of the art object detection network on monocular pedestrian detection;
(2) combining several instances of the early layers of that network into a multi-view deep network whose outer layers are trained for multi-view appearance-based joint detection on relatively smaller multi-camera data-set.

Such architecture allows for processing the input from the separate views in parallel, and the subsequent interconnection layers allow for automatically learning how features across different views map into each other.
Given the trained monocular models, the method we propose is straightforward to reuse as it requires only to retrain on a small multi-camera data-set.
Note that we focus on per-frame processing, while utilising time consistency could be further extension.

In addition, we make public a new medium size three-view data-set of non-occluded pedestrians with fairly more accurate calibration then the largest existing multi-view data-set (PETS 2009).
Our source code and trained models will be made available under an open-source
license.

As a summary, our main contributions are:
(1) proposing the first full deep learning multi-camera people detector;
(2) first empirical results of the superiority of such a method;
as well as
(3) a new three-view data-set with fairly more accurate calibration in terms of consistency of a projection across all of the views.

The monocular methods overview in \S~\ref{sec:related-work}
is of those methods which utilize deep learning,
whereas a complete overview is given for the more 
closely related methods to this work - 
the joint multi-camera people detectors.
The problem on which this paper focuses is formally stated 
in \S~\ref{sec-deepmvpd}, where we also introduce the
multi-view architecture that we  propose.
Our empirical evaluation presented in \S~\ref{sec:experiments} 
demonstrates the encouraging superiority of the multi-view detection
with a deep learning model, and also gives insights about practical
considerations which arise.

\section{Related work}\label{sec:related-work}

\subsection{Deep monocular pedestrian detection} \label{ss-rw_mono}

Applying the R-CNN algorithm~\cite{rcnn} to monocular pedestrian detection~\cite{Hosang2015Cvpr} exceeded the state of the art methods at that time on the Caltech pedestrian data-set~\cite{caltech}, being the first demonstration that CNNs are well-suited to the task.
Later it was shown  that the explicit dealing with pedestrian occlusions, by fine-tuning a separate fully convolutional network for sub-parts of the bounding boxes, provides a 50\% relative improvement~\cite{deepParts2015ICCV}. On top of the part-designated CNNs the authors employ a linear SVM, sparsified to reduce the computational load.

In~\cite{cascades} a Boosting algorithm is proposed, which merges the proposal generation and the detection steps. The cascade learning is formulated as Lagrangian optimisation of a risk accounting for both accuracy and complexity, and integrates both hand-crafted and convolutional features.
The downstream classifier of the extension of R-CNN known as Faster
R-CNN~\cite{fasterrcnn} was recently shown to degrade the performance, and it was proposed to be replaced by boosting forests~\cite{Zhang2016Eccv}.
The resulting method does not use hand-crafted features and reaches state of the art accuracy.

\subsection{Multi-view occupancy map estimation methods}\label{ss-rw_mv}

The first method which uses multi-view streams jointly is the Probabilistic Occupancy Map (POM)~\cite{fleuret-et-al-2008}.
Based on a crude generative model, it estimates the probabilities of occupancy through mean-field inference, naturally handling occlusions.
Further, it can be combined with a convex max-cost flow optimization to leverage time consistency~\cite{pomlp}.
In~\cite{alahi2011sparsity} the problem is re-casted as a linear inverse, regularized by enforcing a sparsity constraint on the occupancy vector.
It uses a dictionary whose atoms approximate multi-view silhouettes.
To elevate the need of O-Lasso computations, in~\cite{scoop} a regression model was derived which includes solely Boolean arithmetic and sustains the sparsity assumption of~\cite{alahi2011sparsity}.
In addition, the iterative method is replaced with a greedy algorithm based on set covering.

In~\cite{peng:mbn} the occlusions are explicitly modeled per view by a
separate Bayesian Network, and a multi-view network is then
constructed by combining them, based on the ground locations and
the geometrical constraints.
Although considering crowd analysis, the multi-view image generation of~\cite{Ge2010} is  with a stochastic generative process of random crowd configurations, and then maximum a posteriori (MAP) estimate is used to find the best fit with respect to the image observations.

It is important to note that  all of these multi-view methods rely on background subtraction pre-processing.

\section{Deep Multi-View People Detection}\label{sec-deepmvpd}

\subsection{Problem definition}\label{sec-def}

\begin{figure}
		\includegraphics[width=\linewidth,trim={2cm 20.5cm 2cm 4.5cm},clip]{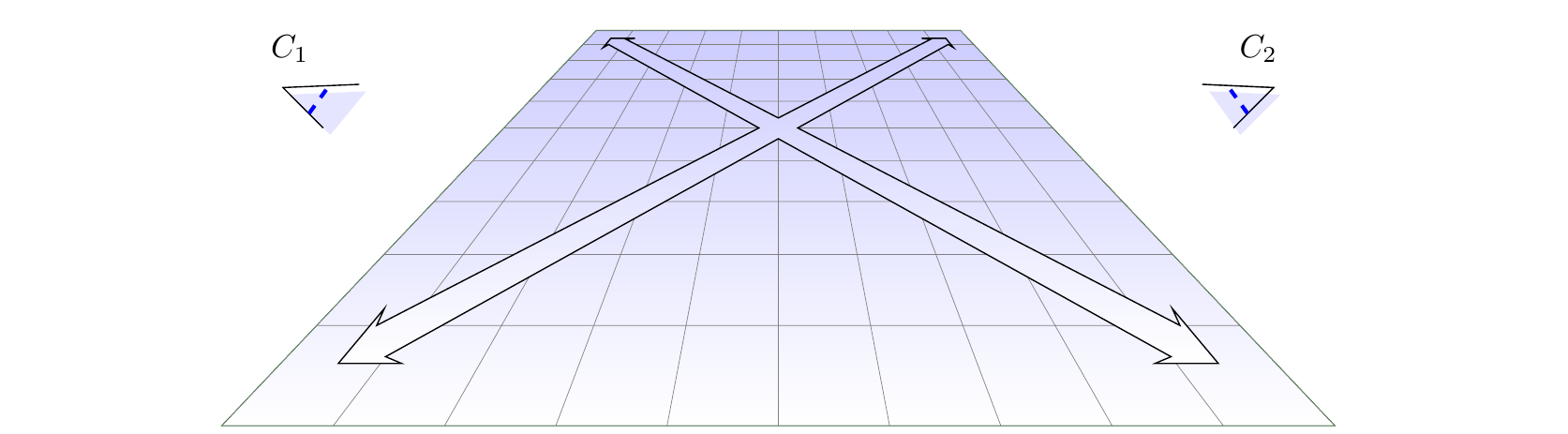} 
	\caption{Illustration of the people occupancy map estimation setup with static cameras.
	As an example we depict two cameras and the area is discretized into a $10\times10$ grid, thus following the notation in the text: $C=2$, $G=100$.
	}\label{fig:mv_ground}
\end{figure}

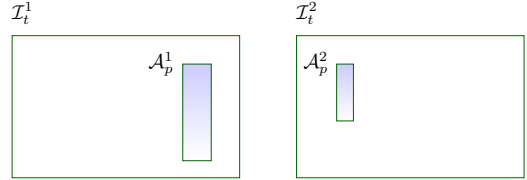
\begin{figure}
\centering
\noindent\resizebox{.8\linewidth}{!}{
\begin{tikzpicture}[object/.style={thin,double,<->}]
	\draw[draw=green!40!black] (-5,-5) rectangle (-1,-2.5);  
	\shade[top color=blue!20, bottom color=white, draw opacity=0.3] (-2,-4.7) rectangle (-1.5,-3);
	\draw[draw=green!40!black] (-2,-4.7) rectangle (-1.5,-3);
	\node[text width=1cm] at (-4.5,-2.1) {${ \cal I}_t^1$};
	\node[text width=1cm] at (-2.1,-3) {${ \cal A}_p^1$};

	\draw[draw=green!40!black] (0,-5) rectangle (4,-2.5);

	\shade[top color=blue!20, bottom color=white, draw opacity=0.3] (0.705,-4) rectangle (1,-3);
	\draw[draw=green!40!black] (0.705,-4) rectangle (1,-3);
	\node[text width=1cm] at (0.5,-2.1) {$ {\cal I}_t^2$};
	\node[text width=1cm] at (0.62,-3) {${ \cal A}_p^2$};

\end{tikzpicture}
}
\caption{The input to the model when estimating $q_p$ consists of 
the cropped regions $\textbf{A}_{{p}}$ of $\textbf{I}_t$ (see \S~\ref{sec-def}).
The illustrated example is in-line with the one in Fig.~\ref{fig:mv_ground}.
}
\label{fig:input}
\end{figure}

We discretize the area common to the fields of view of $C$ cameras in a regular grid of $G$ points or interchangeably - cells, as illustrated in Fig.~\ref{fig:mv_ground}.
To estimate the occupancy of a cell $p$, we consider a cylinder centred 
at the $p$th position, whose height corresponds to 
the one of the humans' average height. 
We use the cameras' calibration to obtain the cylinder's projections into the views where it is visible. These rectangular projections yield the cropped regions $\textbf{A}_{{p}}$ of $\textbf{I}_t$, 
as illustrated in Fig. \ref{fig:input}.

At each time step $t$ we are given a set of images $\textbf{I}_t = {\cal I}^1_t, \dots {\cal I}^C_t$ taken synchronously. 
Hence, given sub-images cropped for a particular position ${{p}}= 1, \dots, G$ 
in the separate views
$\textbf{A}_{{p}} = \left( {\cal A}^1_{{p}}, \dots, {\cal A}^C_{{p}} \right)$,
we aim at estimating the probability of it being occupied, that is
\[
q_{{p}} = p(X_{{p}} = 1 \mid \textbf{A}_{{p}}),
\]
where $X_{{p}}$ on $\{ 0, 1 \}$ stands for the position ${{p}}$ being free and occupied, respectively. In the following, let ${\cal I}$ be the set of all possible cropped sub-images.

\subsection{Method}

Ideally, given large scale multi-camera data-set, one could train a complete multi-stream processing model.
However, due to the data-set problem discussed above as well as the fact that in practice the annotated data is often scarce, the method we propose consists of multiple steps,
so as to take advantage of the existing larger scale monocular pedestrian data-set Caltech~\cite{caltech} and to generalize better.
We separately elaborate these steps.

\subsubsection{Monocular fine-tuning with input-dropout}\label{sec:monoc-fine-tuning}
Primarily, in order to learn discriminative features of the pedestrian class, a monocular pedestrian detecting CNN has to be trained.
To ease the training procedure we recommend starting from a pre-trained object detection CNN.
In particular, in our experiments we started from either GoogLeNet~\cite{inception2015} or AlexNet~\cite{alexnet2012}, trained on ImageNet~\cite{imagenet_cvpr09}
and we replaced the last fully connected layer with a randomly initialized one with two output units.
We then fine-tune the resulting network on the Caltech data-set (see \S~\ref{ss-data-sets}).
As stated in the previous sections, our training samples take the form of rectangular sub-images cropped in one of the full image.

\begin{figure}[!t]
  \centering
  \includegraphics[width=\linewidth,trim={0.7cm 15.3cm 12.1cm 0.73cm},clip]{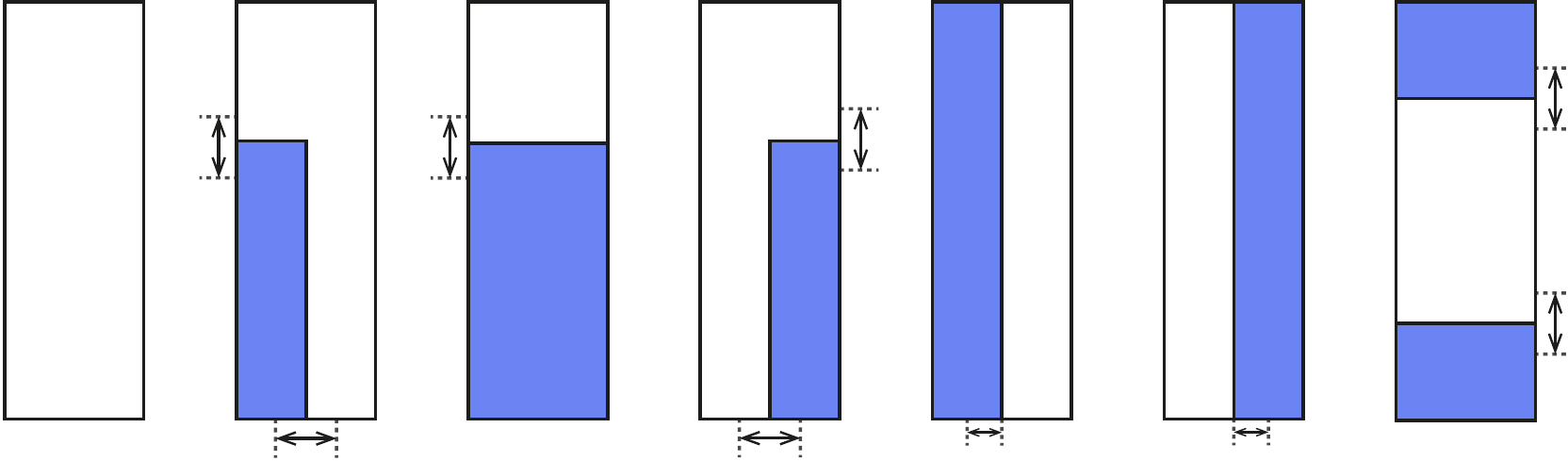}
  \caption{Occlusion masks applied to the input, illustrated with shaded area.
  Among these, the first represents no occlusion.
  As the arrows indicate, their width and/or height are restrictively randomized within a margin.}
  \label{fig-masks}
\end{figure}

Experimental results on monocular pedestrian detection indicate that state of the art detection accuracy is achieved by explicitly dealing with the occlusions (for example by detecting parts of body), as explained in \S~\ref{ss-rw_mono}.
As we shall see in the following sections, the multi-view architecture finally contains components initialized with the weights of the monocular detector. It is necessary that the occlusion awareness of the model is implemented in an efficient manner.
Thus, to improve the robustness to occlusion, we propose a novel technique dubbed \textit{input dropout}, which consists in augmenting the input data by artificially masking part of the signal. While the masking is adapted to the morphology of the positive class, it has to be done on both positive and negative samples during training, so that it does not provide erroneous discriminating cues to the network.

In our experiments in particular we defined $7$ rectangular occluding masks--\#1 being ``no occlusion at all''--(Fig.~\ref{fig-masks}), and for each sample, negative or positive, we pick one of the masks uniformly at random, and replace the pixel it masks with white noise. As for drop-out, this ``input drop-out'' forces the network toward greater redundancy between representations.

\subsubsection{Multi-camera architecture}\label{sec:multi-camera-comb}

\begin{figure}
		\includegraphics[width=\linewidth,trim={1.5cm 14.2cm 4cm 4.3cm},clip]{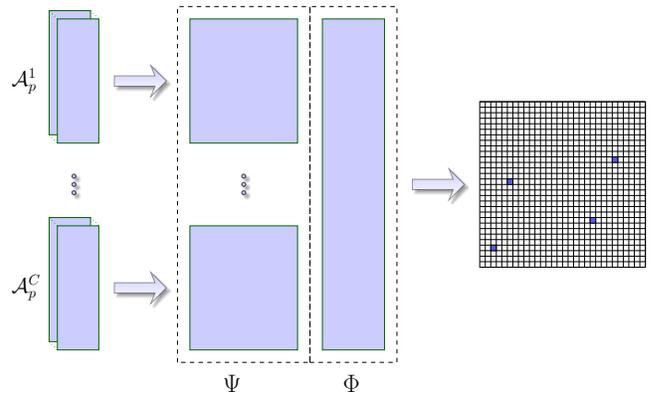} 
	\caption{Illustration of the multi-view architecture. On the left-most side is represented the input; followed by the multi-view architecture consisting of two conceptual parts $\Psi$ whose components process the input in parallel, and of $\Phi$; and on the right most side a top-view of the predictions is illustrated where a coloured cell represents a detection.
	}\label{fig:mv_arch}
\end{figure}

Given a CNN trained for monocular detection, we retain $d$ of its layers, resulting in an embedding $\psi: {\cal I} \rightarrow \mathbb{R}^Q$, where $Q$ is the number of output units of the $d$th layer. The concatenation of $C$ such embeddings is a multi-view embedding ${{\Psi}}: {\cal I}^C \rightarrow \mathbb{R}^{CQ}$, on top of which we train a binary classifier $\Phi: \mathbb{R}^{{CQ}} \rightarrow \mathbb{R}^2$, as illustrated in Fig.~\ref{fig:mv_arch}.

\textbf{On the choice of type of classifier of $\Phi$.}
Particular object seen from different perspectives/cameras could possess both features which are consistent across views: \latin{e.g.} color; and some which are different: for example a line curved to the left would be curved to the right viewed from the opposite side. Note that the former is more likely when the two cameras are quite close to each other, and the latter in the opposite case.

The goal of $\Phi$ is to learn how features from different views map to each other.
Thus, a natural choice for $\phi$ is a  multi-layer perceptron (MLP), in aim to "wire together what fires together".
In our experiments the output of ${{\Psi}}$ is flattened, and ${{\Phi}}$ is a MLP, and for demonstrative experimental analysis $\Phi$ is also a (forest of) decision tree(s).

\textbf{On the choice of \textit{full fine-tuning} or training only $\Phi$.}
To allow the embedding to capture cues discarded for monocular detection but which are informative for multi-view, the complete multi-stream architecture, that is both $\Psi$ and $\Phi$, can be trained, when the latter is differentiable.

On the other hand, keeping $\Psi$ fixed makes the capacity of the predictor family low, and hence prevents over-fitting when the training multi-view data-set is small,
\latin{e.g.} in the order of few hundreds of examples.
Given the size of the data-sets we have used, this approach performed worse than keeping $\Psi$ fixed.

\subsubsection{Non Maxima Supression}\label{sec:nms}
Finally, we perform a non-maxima suppression (NMS) of the detection candidates for all the positions projected in one of the views.
This step selects the final detections so that out of those candidates overlapping more then a predefined threshold only one remains.
Differently then standard NMS implementations, we take into account the detections' scores, in such a way that the priority of a detection candidate to be selected is proportional to its detection score.

\section{Experiments}\label{sec:experiments}

\subsection{Data-sets and metrics}\label{ss-data-sets}

\textbf{Caltech10x}~\cite{caltech}.
For monocular training, we use the Caltech-USA pedestrian data-set and its associated benchmarking toolbox~\cite{PMT}.
This set consists of fully annotated 30 Hz videos taken from a moving vehicle in a regular traffic, and as in other recent works, we use 10 fps sampling.
This increases the training data to $\simeq 2 \cdot 10^4$ detections.
We generate proposals with the \textit{SquaresChnFtrs} detector~\cite{Benenson2015}, and use as positive examples the proposals with $0.5$ overlap threshold.

\vspace*{0.5em}

\noindent \textbf{PETS 2009 S1 L2 data-set}~\cite{pets}.
For comparison with existing methods we use the PETS 2009 data-set, which is a sequence of 795 frames recorded with seven static outdoor cameras.
As noted by other authors:~\cite[p. 10]{peng:mbn}, or~\cite[p. 10]{Ge2010},
we also observe inconsistencies  both in terms of calibration and synchronization.

\vspace*{0.5em}

\begin{figure*}[!htb]
	\begin{minipage}{.32\linewidth}
			\includegraphics[width=\linewidth,trim={2cm 0cm 1cm 0cm},clip]{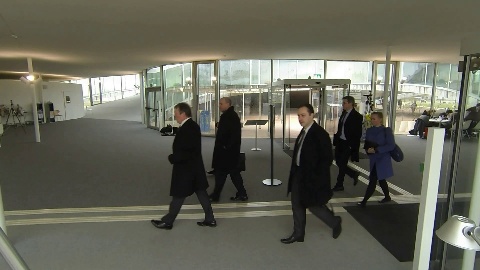} 
	\end{minipage}\hfill
	\begin{minipage}{.32\linewidth}
		\includegraphics[width=\linewidth,trim={3cm 0cm 0cm 0cm},clip]{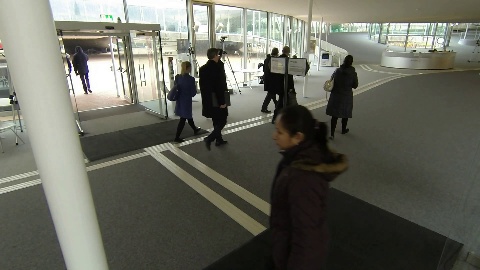} 
	\end{minipage}\hfill
	\begin{minipage}{.32\linewidth}
		\includegraphics[width=\linewidth,trim={0cm 0cm 3cm 0cm},clip]{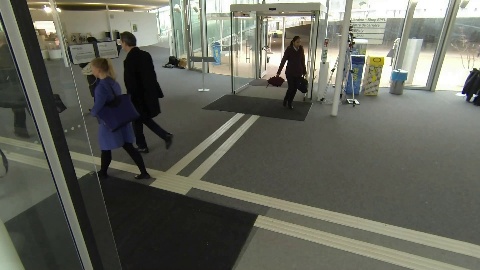}
	\end{minipage}\hfill
	\caption{Example of synchronized corresponding frames of the three views of the new EPFL-RLC data-set.}\label{fig:epfl-rlc}
\end{figure*}

\noindent \textbf{EPFL-RLC data-set}~\cite{RLCdataset}.
This new corpus was captured
with three calibrated HD cameras, with a frame rate of  $60$ frames per second.
An example of across-views corresponding images at a particular time is illustrated in Fig.~\ref{fig:epfl-rlc}.
Currently the annotations represent a balanced set of $4088$ multi-view examples.
Note that a negative multi-camera example may contain a pedestrian in one of its views.
Thus, for each view the negative samples contain a bit of annotated information wheter it contains a pedestrian or not. This allows the data-set to be used for monocular pedestrian training in which case its size is increased three-fold.

Full ground-truth annotations are provided for the last $300$ frames of the sequence, intended to be used for testing while ensuring diversity of the appearance with respect to the training data.
As for future work, we also make available the full sequence of $8000$ synchronized frames of each view.

\vspace*{1em}

\noindent \textbf{Metrics}
Apart from standard classification measures, we use the Multiple Object Detection Accuracy (MODA) metric, accounting for the normalized missed detections and false positives, as well as the Multiple Object Detection Precision (MODP) metric, which assesses the localization quality of the true positives~\cite{metric}.
We also estimate the empirical precision and recall of the detector,
calculated by $P=TP/{TP+FP}$ and $R=TP/{TP+FN}$ respectively,
where: $TP$, $FP$, and $FN$ are the true positives, false positives and
the false negatives, respectively.

\subsection{Monocular pedestrian detection}

\begin{figure}[tp]
		\includegraphics[width=\linewidth,trim={0.5cm 0cm 1cm 1.1cm},clip]{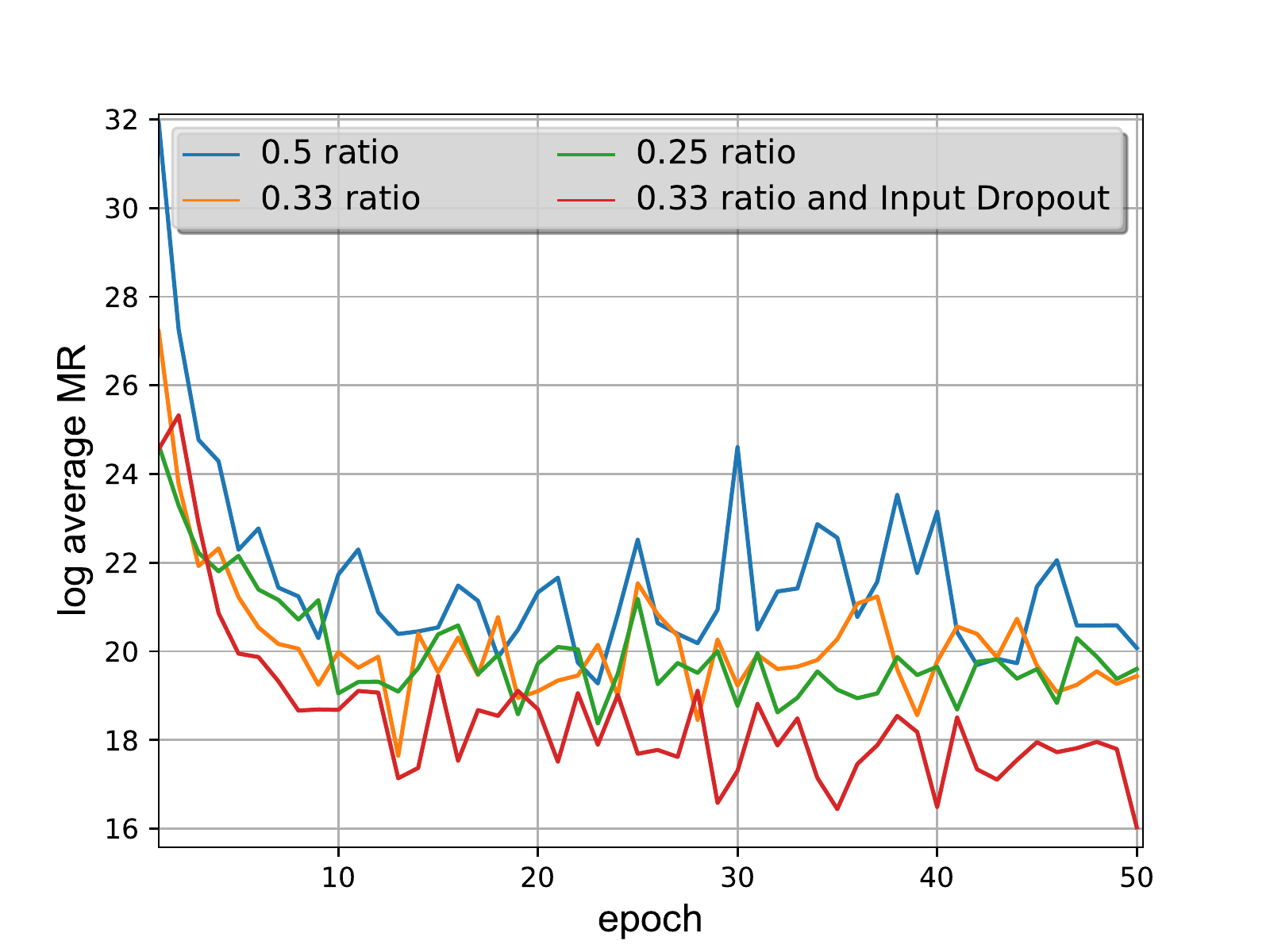}
	\caption{Log average MR on monocular training of GoogLeNet, while using different ratios of positive and negative samples per batch, as well as when using input dropout, as explained in \S~\ref{sec:monoc-fine-tuning}.}\label{fig:goo_mono}
\end{figure}

We conducted our experiments using the Torch framework ~\cite{torch}, and mainly with GoogLeNet~\cite{inception2015} and AlexNet~\cite{alexnet2012}.
The input dimension for these models are respectively $224\times224$ and $227\times227$.
In monocular detection, we observed that training with square input marginally outperforms warping.
In particular, as a performance measure we use the log average miss rate (MR),
as it is widely adopted in monocular pedestrian detection.

In Fig.~\ref{fig:goo_mono} we illustrate an example experiment in which we used GoogLeNet, batch size of $64$, learning rate $0.005$, SGD with momentum $0.9$, as well as a proportion of positive samples per mini-batch of $r=0.33$. We observe mean MR of $19.81$ ($\sigma=0.58$) on the test data when the learning starts to saturate.
When input dropout is applied the mean MR drops to $17.32$ ($\sigma=0.75$), and further continues to decrease, although with slight variance. 
Applying it allowed for reaching MR performances which were never reached with standard training, and the observed variance suggests that combining it with more sophisticated learning rate policy
can provide further performance gain.

With AlexNet, the mean MR is $24.04$ ($\sigma=0.39$), and $22.43$ ($\sigma=0.51$), without and with input dropout, respectively.
Although marginal, input-dropout demonstrated consistent improvement in our experiments.

As a summary of our experiments, we consistently observe performance improvement when using:
(1) GoogLeNet instead of AlexNet,
(2) square cropping versus warping the region,
(3) a proportion of $r=0.33$ positive samples per mini-batch, and
(4) when input dropout (\S~\ref{sec:monoc-fine-tuning}) is applied.
By doing this, we reduced the MR to 15.61\%, which is higher than the results reported in ~\cite{deepParts2015ICCV}, but requires six times less computation and is simpler to implement and deploy.

\subsection{Multi-camera people detection}
In our experiments, we discretize the ground surface of the EPFL-RLC and PETS data-sets to grids of $45\times55$ and $140\times140$ positions, respectively.
To train the multi-stream model on the PETS data-set we automatically extract negative examples which outnumber the positive ones provided by the annotated detections. During one epoch of training, these negatives are sampled without replacement. Both on EPFL-RLC and PETS data-sets we observe improved performances when training with forced ratio of increased negative samples per mini-batch.

Contrary to the monocular training, we observed no performance drop of training $\Phi$ if rectangular input is being used, which is why we extract features with reduced input in the width dimension by omitting $50$ pixels of both sides. In our implementation, we zero-pad the input of the view for which a particular position is out of its field of view.

Unless otherwise stated, we use MLP with $3$ fully connected layers, ReLU non-linearities and the output is re-scaled using log-softmax.
Natural consideration is the use of regularization as intuitively the weight vector of the fully connected layers would be sparse.
We experimented with the $p$-norm regularization term:
$\|w\|_p = (\sum_{i=1}^N |w_i|^p)^{1/p}$, both with $p=1,2$, while using SGD optimization, but did not observe a substantial improvement.
In fact, by visualizing the weights when training without regularization we observe that major part of the weight vector tends towards zero, and that the training was more stable.

We provide bellow an empirical analysis of the impact on performance of some aspects of our setup.

\subsubsection{Training with hard-negative auto-generated examples}
Since the EPFL-RLC data-set demonstrates more accurate joint-view calibration then PETS'09 we are able to automatically generate multi-view negative samples which we refer to as \textit{hard negatives} and which  force $\Phi$ to compare the multi-stream extracted features in a way that generalizes better.

We perform this  automatically in two different ways:
(1) by shifting the rectangles of a positive detection in one or two of the views, or
(2) by combining positive detections of different persons.

\begin{figure}[!b]
  \minipage{.495\linewidth}
  \includegraphics[width=\linewidth,trim={.2cm 0.1cm .1cm 0.1cm},clip]{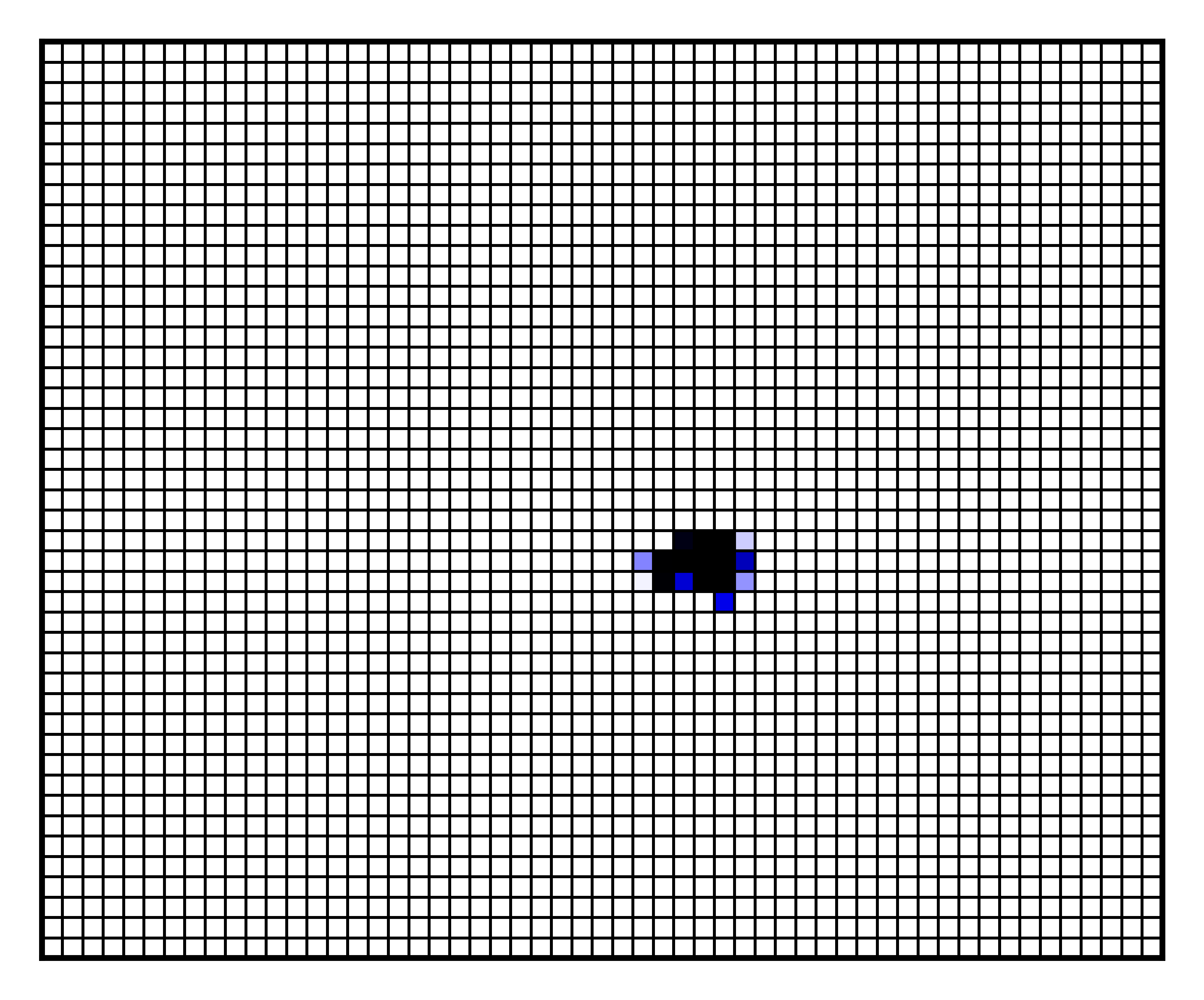}
  \endminipage\hfill
  \minipage{.495\linewidth}
  \includegraphics[width=\linewidth,trim={.1cm 0.1cm .2cm 0.1cm},clip]{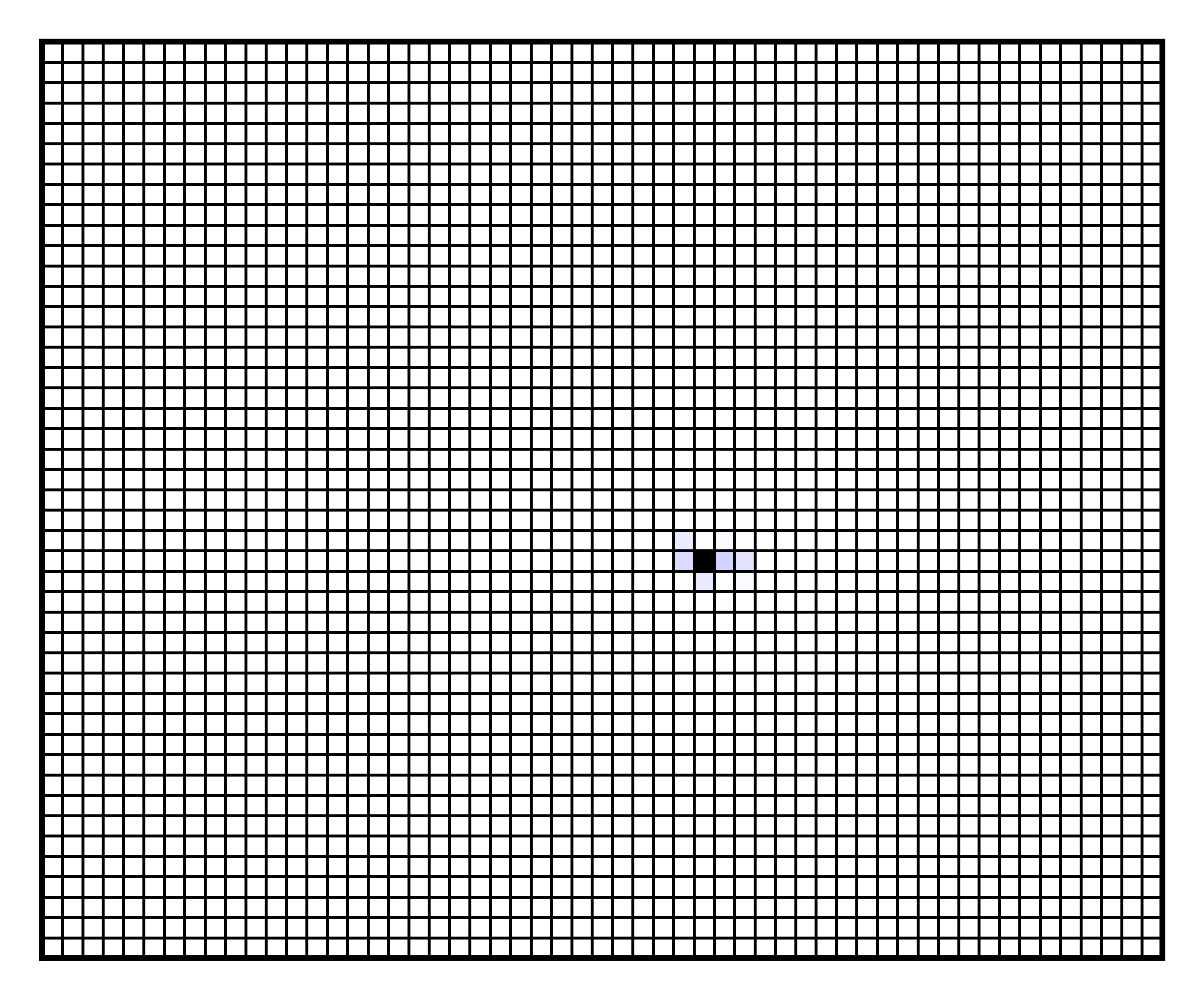}
  \endminipage\hfill
  \caption{Top view visualization of the row probability outputs on a randomly selected frame of the EPFL-RLC data-set. Left: training without hard negatives; right: using hard negatives (\S~\ref{ss-data-sets}).}
  \label{fig-topViewProbas}
\end{figure}

Training with these hard negatives allows respectively for:
(1) sharper detections in the ground plan, which makes the method less sensitive to the non-maxima suppression post-processing as shown in Fig.~\ref{fig-topViewProbas}; and
(2) forcing the classifier $\Phi$ to learn a stronger joint appearance model.
The former improve localization and the latter to decrease the false positives of the detector.
Note that in the experimental results presented in the following, changing the ratio $r$ implies changing the difficulty of the classification for this data-set and as we shall see hurts the accuracy.

\begin{figure*}[!htb]
	\includegraphics[width=\linewidth]{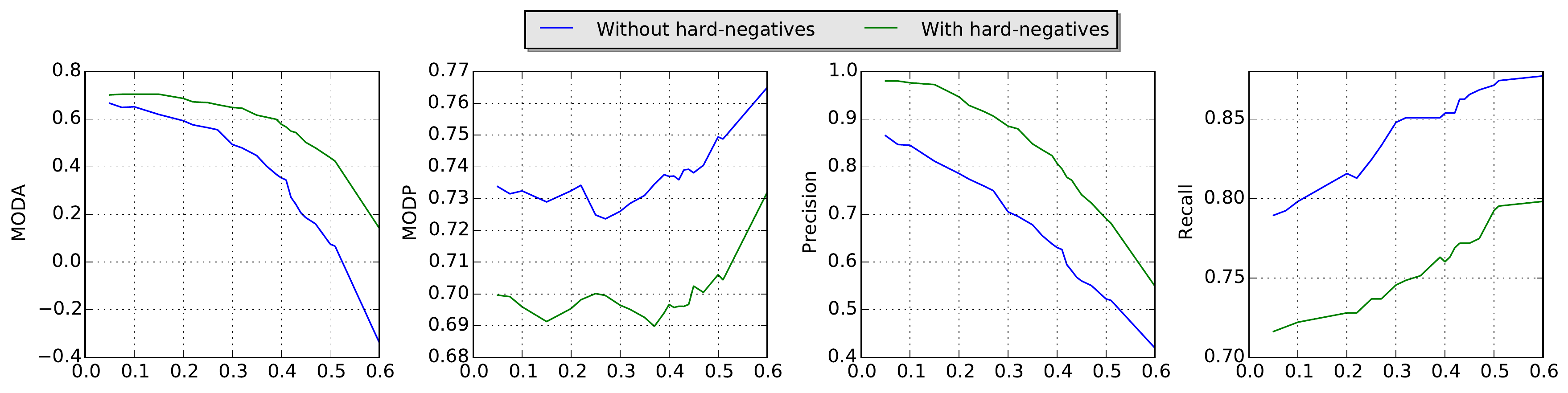}
	\caption{Comparison of training with and without hard multi-view negative examples, versus the NMS threshold (x-axis) on the EPFL-RLC data-set.}\label{fig-hard_neg}
\end{figure*}

From the results illustrated in Fig.~\ref{fig-hard_neg} we observe that training with such negatives improves the most important MODA metric and reaches almost perfect precision, at the cost of marginally decreasing the MODP metric, and decreased recall due to the slightly increased number of missed detections.
However, as the camera calibration of the PETS data-set demonstrates non-negligible
joint-camera inconsistency of the projections,
we did not experiment with the multi-view hard negatives for it, to avoid increasing the noise level of the input.

\subsubsection{Monocular Vs. Multi-view classification}

We compare performance with and without using multiple views.
As the performance using a MLP for $\Phi$ is potentially highly dependent on the persistence of tuning the hyper-parameters and/or the choice of optimization methods,
in this section we use solely random forests for $\Phi$.

\begin{figure*}[!tbp]
  \noindent\begin{minipage}{.48\textwidth}

  \centering
  \subfloat[]{\includegraphics[width=\linewidth,trim={1.2cm 0.8cm 2.05cm 2.1cm},clip]{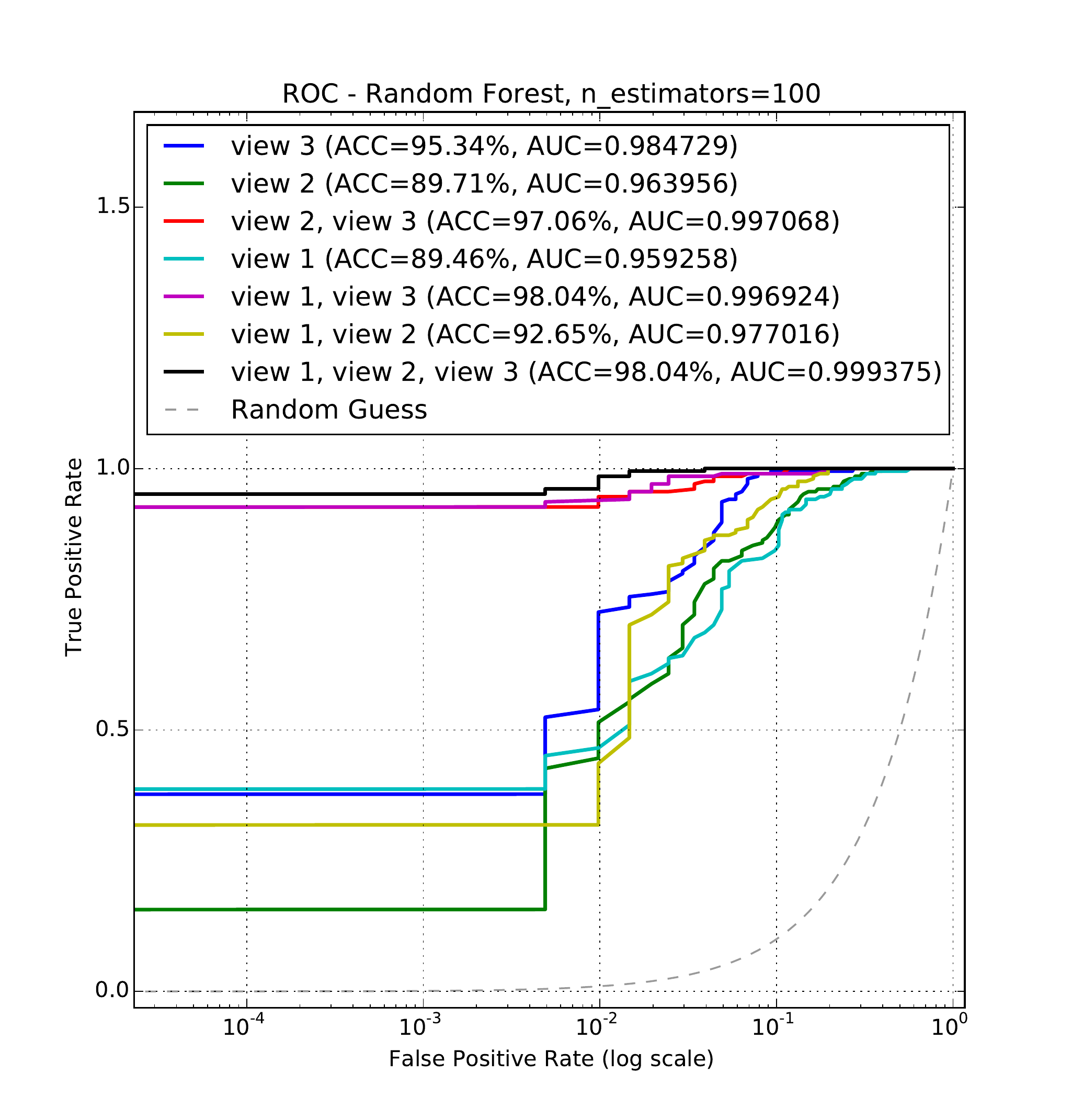}\label{fig-rf_epfl-1}}

  \end{minipage}
  \noindent\hspace{0.5cm}\begin{minipage}{.48\textwidth}
  \subfloat[]{\includegraphics[width=\linewidth,trim={1.2cm 0.8cm 2.05cm 2.1cm},clip]{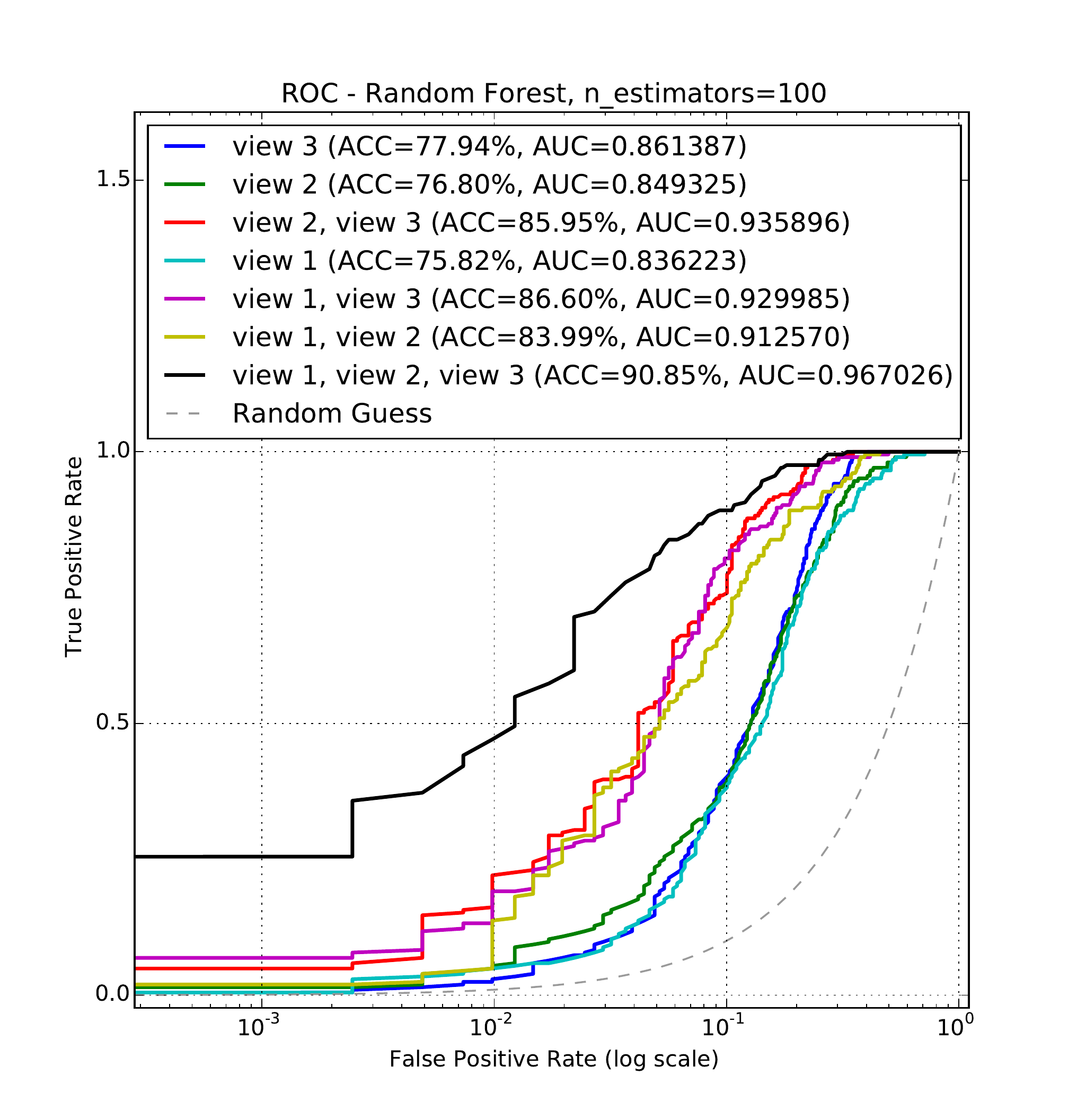}\label{fig-rf_epfl-2}}
  \end{minipage}
  \caption{Accuracy, ROC and AUC on the EPFL-RLC test data, using random forest of 100 trees, with $d=23$ and GoogLeNet, as well as ratios: \protect\subref{fig-rf_epfl-1} $r=0.5$ in which case we only use easy negative samples, and \protect\subref{fig-rf_epfl-2} $r=0.33$.
  }\label{fig-rf_epfl_nviews}

\end{figure*}

For this purpose, we use the accuracy and the ROC while considering all possible (sub)sets of the available views, both on PETS and EPFL-RLC, as well as for different values of the proportion $r$ of positive samples. In all experiments, we consistently observe performance improvement as more views are being added, as illustrated on Fig.~\ref{fig-rf_epfl_nviews} for the EPFL-RLC data-set. In particular, we see from Fig.~\ref{fig-rf_epfl_nviews} that it is either the case that the accuracy is improved or if it is already good the confidence of the classifier increases significantly.

\subsubsection{Impact of the Selected Layer on the Accuracy}\label{ss-layer}
We use GoogLeNet and compare using the first 11, 15, 20 and 23 layers.
After flattening the output of each of the layers, we add fully connected layers, each containing 1024, 512 and 2 output units.
We experiment with:
(1) mini-batch size (32 and 64),
(2) optimisation technique: SGD w/o momentum, Adam~\cite{adam}, Adadelta~\cite{adadelta} and RMSPRop~\cite{rmsprop}; as well as
(3) different initial values for the learning rate when necessary.
We run each experiment for 60 epochs on the EPFL-RLC data-set, and we list in Table~\ref{t_layer_dep_epfl} the best accuracy at that point for each of the selected layers.
In the majority of the experiments we observe best performances with Adadelta or ADAM, and batch size of 64.

Our experiments indicate that the depth of the CNN used as feature extractor
does not impact much the performance.
This is encouraging since shallower structures can be used what decreases 
the computation time when the trained model is used.

\begin{table}[!b]
\begin{center}
\caption{Test accuracy on the EPFL-RLC data-set, for several numbers of
  convolution layers $d$, and  proportion $r$ of positive
  samples per batch (see
  \S~\ref{sec:multi-camera-comb}) with different level of difficulty (see \S~\ref{ss-data-sets}).}\label{t_layer_dep_epfl}
\vspace*{0.5em}

\begin{tabular}{l|cccc}
\toprule
\diagbox[width=4em,height=1.9em]{$r$}{$d$} & 11 & 15  & 20  & 23  \\
\toprule
0.50          & 99.75\% & 100.00\% & 99.75\% & 98.77\% \\
0.33          & 96.24\% & 97.05\%  & 96.73\% & 94.60\% \\
0.25          & 91.78\% & 83.57\%  & 83.70\% & 81.74\% \\
\bottomrule
\end{tabular}
\end{center}
\end{table}

\subsubsection{Multi-view Information}\label{s-mutal_info}
We now move our focus on the question whether the multi-view joint method
provides information which allows for easier discrimination among the two classes,
compared to the case of using solely one view.
To do so, we analyze if features from the different views are used uniformly, or if features from one of the views dominate the decision.

The easiest way to illustrate this is by using a single decision tree for $\Phi$ and investigating the distribution of the most important features across the views.
Legitimately, we define the importance of a feature in terms of its depth in the decision tree: the higher, or the closest to the root, the more important a feature is.

\begin{figure}[!t]
\centering
\includegraphics[width=\linewidth, trim={0.5cm 0.0cm 1.3cm 1.05cm},clip]{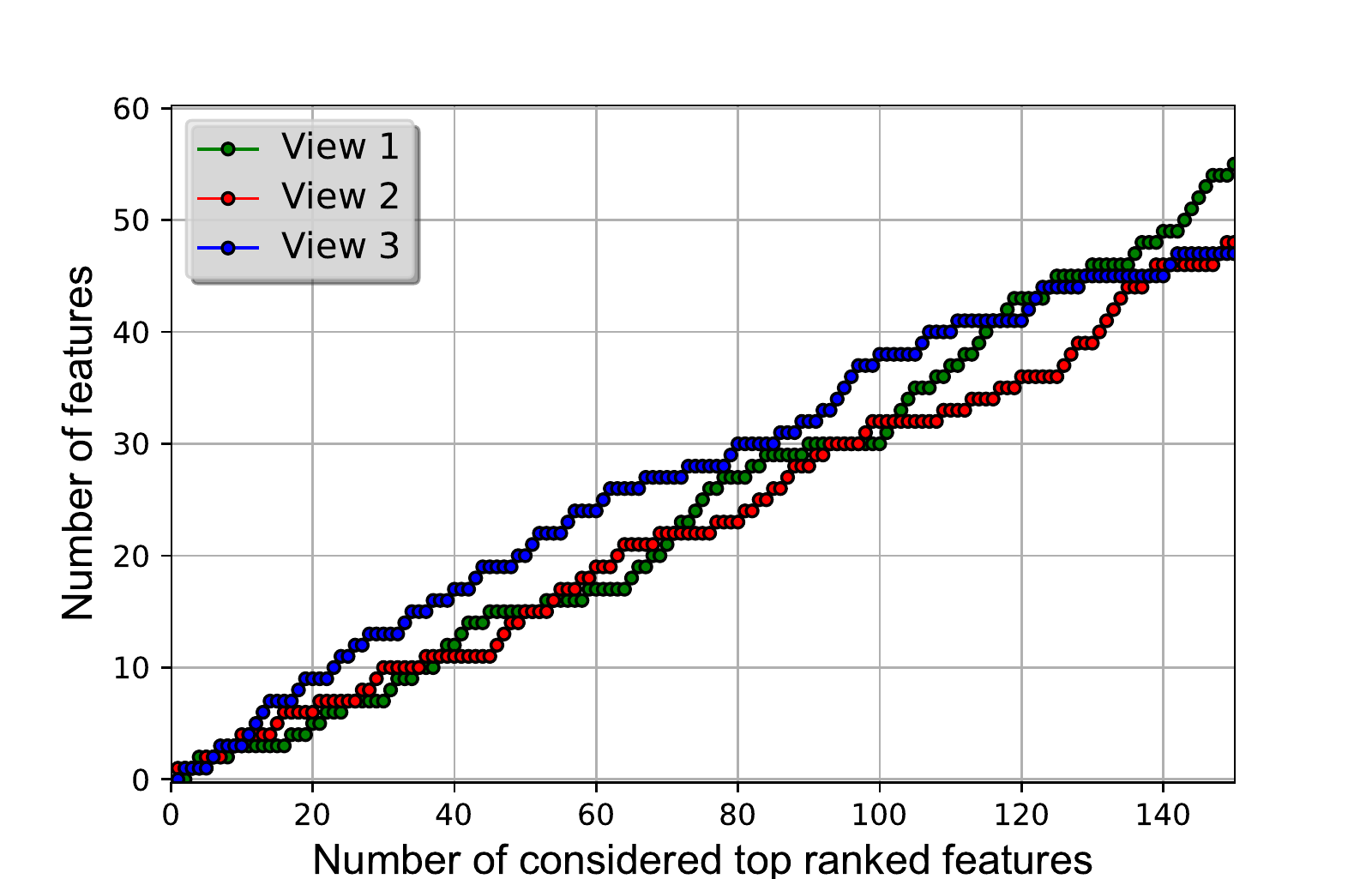}
\caption{Distribution of the most important features across the views.
For details please refer to \S~\ref{s-mutal_info}.}
\label{fig_topftrs}
\end{figure}

In the results illustrated in Fig.~\ref{fig_topftrs} we use the EPFL-RLC data-set and GoogLeNet with $d=23$, which for the three views provides us with $3 \times 1024$ features. We visit the nodes from the root to the leaves, and count the features from each view. We find that among the $50$, $100$ and $150$ top-nodes, the numbers of features from each view are $16/15/19$, $36/32/32$, and $55/51/44$, respectively. This shows that the classifier exploits the views in quite a balanced manner.

\subsection{Comparison to existing methods on PETS}\label{sec:comp-exist-meth}
Finally, we compare our method with the current state of the art multi-view methods.
In Table~\ref{t_mvcmp}, we call ``ours'' the experiments where we use the full PETS sequence as testing, and the model has never being trained on it, and we call ``ours-ft''  the experiments where we divide the sequence to training for fine-tuning the model, and testing.
In the latter we consider all the possible splits to train and test frames, and we list the mean and the standard deviation of the particular performance measure estimated through these multiple folds.

The experimental results justify the multiple steps of our approach.
As the results in Table~\ref{t_mvcmp} show, without data normalization
specific to this dataset and without fine-tuning we outperform 
the existing methods, what indicates generalization property 
which is of interest.

Further, we recall the calibration inconsistencies of the PETS dataset which become more prominent as we use more of the views of this dataset. Authors address this deterioration of the
homograph mapping in the presence of a slope in the scene \cite{peng:mbn},
causing degradation in performance when adding particular views.
To the best of our knowledge researchers avoid utilizing the seven views together.
Instead, most common is the case of using solely the first view;
and in other cases, authors utilize a subset of the views. 
Of the latter, most often reported is the case of using the views $1,4,5$ and $7$ (or $1,5,6,8$ in the original enumeration).
It is interesting that in our experiments we do not observe 
performance drop when all of the seven views are used, 
and that in fact we observe marginal improvement.
This could be due to the contextual padding 
that we recommended earlier.
Further, the fact that the MODP metric increases, 
may indicate that $\Phi$ is able to learn these inconsistencies.

\begin{table}[!t]
\caption{Comparison on PETS S2L1, using either a single view, or using
  our multi-view approach (see \S~\ref{sec:comp-exist-meth}). 
  Note that the views' enumeration omits the second view 
  which was removed from the dataset.}\label{t_mvcmp}
\setlength{\tabcolsep}{5pt}
\renewcommand{\arraystretch}{1.1}
\begin{center}

\begin{tabular}{cccccc}
\toprule
Method    		& Views      & MODA        & MODP        & Precision   & Recall      \\
\hline
\cite{peng:mbn}  & 1,4,5,7	& 0.79        	& 0.74        	& 0.92        	& 0.91        	\\
\cite{Ge2010}    & 1,4,5,7  & 0.75        	& 0.68        	& 0.85        	& 0.89        	\\
Ours-ft          & 1		& 0.94 (0.03) 	& 0.61 (0.02) 	& 0.97 (0.02) 	& 0.96 (0.02) 	\\
Ours             & 1		& 0.88        	& 0.75        	& 0.97        	& 0.91        	\\
Ours-ft       	& 1,4,5,7  	& 0.90 (0.05) 	& 0.66 (0.02) 	& 0.94 (0.03) & 0.96 (0.02) \\
Ours-ft       & 1-7  & 0.94 (0.02) & 0.68 (0.02) & 0.98 (0.01) & 0.97 (0.02) \\
\bottomrule
\end{tabular}

\end{center}
\end{table}

As shown, our method outperforms the existing state of the art baselines, which is in large part due to its ability to leverage appearance, and filter out non-human foreground objects that may trigger a background-subtraction procedure.

\section{Conclusion and Future Work}
We proposed end-to-end deep learning method for multi-view people detection
which outperforms existing methods.

The training consists of two-steps:
(1) training an occlusion-aware model for monocular pedestrian detection;
(2) training a multi-view architecture partially initialized with the monocular one on smaller multi-view data-sets.
The advantage of the latter is processing in parallel the input of the separate views and yielding a decision which jointly exploits the reliable appearance cues.

We demonstrated that:
(1) multi-view classification increases the accuracy and the confidence of the classifier over the monocular case, and
(2) appearance features extracted jointly from multiple views provide more easily separable embedding that in turn allows for more accurate classification than the monocular one.
In addition we discussed various implementation insights.

The presented experimental results motivate promising future work.
Various extensions  can be done which may require larger scale data-set,
both for training and for more challenging evaluation needed to compare 
the methods.
For example, the following directions could be explored:
(1) more explicit multi-view occlusion reasoning;
(2) training domain-adaptation modules.
The former would allow for merging the two steps of generating detection candidates and selecting them, whereas the latter would allow for easier adaptation for the task at hand
sustaining valuable multi-view features automatically captured on such large scale data-set.

Nevertheless, given that in practice it is often the case that the annotated data is scarce,
the method discussed here would probably be more likely to be applied as it provides good accuracy.

\section*{Acknowledgment}

This work was supported by the Swiss National Science Foundation, under the grant CRSII2-147693 "WILDTRACK". We also gratefully acknowledge NVIDIA's
support through their academic GPU grant program.

\bibliographystyle{IEEEtran}
\bibliography{deepmcd}

\begin{thebibliography}{10}
\providecommand{\url}[1]{#1}
\csname url@samestyle\endcsname
\providecommand{\newblock}{\relax}
\providecommand{\bibinfo}[2]{#2}
\providecommand{\BIBentrySTDinterwordspacing}{\spaceskip=0pt\relax}
\providecommand{\BIBentryALTinterwordstretchfactor}{4}
\providecommand{\BIBentryALTinterwordspacing}{\spaceskip=\fontdimen2\font plus
\BIBentryALTinterwordstretchfactor\fontdimen3\font minus
  \fontdimen4\font\relax}
\providecommand{\BIBforeignlanguage}[2]{{%
\expandafter\ifx\csname l@#1\endcsname\relax
\typeout{** WARNING: IEEEtran.bst: No hyphenation pattern has been}%
\typeout{** loaded for the language `#1'. Using the pattern for}%
\typeout{** the default language instead.}%
\else
\language=\csname l@#1\endcsname
\fi
#2}}
\providecommand{\BIBdecl}{\relax}
\BIBdecl

\bibitem{rcnn}
\BIBentryALTinterwordspacing
R.~B. Girshick, J.~Donahue, T.~Darrell, and J.~Malik, ``Rich feature
  hierarchies for accurate object detection and semantic segmentation,''
  \emph{CoRR}, vol. abs/1311.2524, 2013. [Online]. Available:
  \url{http://arxiv.org/abs/1311.2524}
\BIBentrySTDinterwordspacing

\bibitem{Hosang2015Cvpr}
J.~Hosang, M.~O., R.~Benenson, and B.~Schiele, ``Taking a deeper look at
  pedestrians,'' in \emph{Conference on Computer Vision and Pattern
  Recognition}, 2015.

\bibitem{caltech}
P.~Dollar, C.~Wojek, B.~Schiele, and P.~Perona, ``Pedestrian detection: A
  benchmark,'' in \emph{Conference on Computer Vision and Pattern Recognition},
  2009, pp. 304--311.

\bibitem{deepParts2015ICCV}
Y.~Tian, P.~Luo, X.~Wang, and X.~Tang, ``Deep learning strong parts for
  pedestrian detection,'' in \emph{International Conference on Computer
  Vision}, 2015, pp. 1904--1912.

\bibitem{cascades}
\BIBentryALTinterwordspacing
Z.~Cai, M.~J. Saberian, and N.~Vasconcelos, ``Learning complexity-aware
  cascades for deep pedestrian detection,'' \emph{CoRR}, vol. abs/1507.05348,
  2015. [Online]. Available: \url{http://arxiv.org/abs/1507.05348}
\BIBentrySTDinterwordspacing

\bibitem{fasterrcnn}
S.~Ren, K.~He, R.~Girshick, and J.~Sun, ``Faster {R-CNN}: Towards real-time
  object detection with region proposal networks,'' in \emph{Advances in Neural
  Information Processing Systems ({NIPS})}, 2015.

\bibitem{Zhang2016Eccv}
L.~Zhang, L.~Lin, X.~Liang, and K.~He, ``Is faster r-cnn doing well for
  pedestrian detection?'' in \emph{European Conference on Computer Vision},
  2016.

\bibitem{fleuret-et-al-2008}
\BIBentryALTinterwordspacing
F.~Fleuret, J.~Berclaz, R.~Lengagne, and P.~Fua, ``Multi-camera people tracking
  with a probabilistic occupancy map,'' \emph{IEEE Transactions on Pattern
  Analysis and Machine Intelligence}, vol.~30, no.~2, pp. 267--282, 2008.
  [Online]. Available:
  \url{http://fleuret.org/papers/fleuret-et-al-tpami2008.pdf}
\BIBentrySTDinterwordspacing

\bibitem{pomlp}
J.~Berclaz, F.~Fleuret, and P.~Fua, ``Multiple object tracking using flow
  linear programming,'' Idiap, Idiap-RR Idiap-RR-10-2009, 6 2009.

\bibitem{alahi2011sparsity}
A.~Alahi, L.~Jacques, Y.~Boursier, and P.~Vandergheynst, ``Sparsity driven
  people localization with a heterogeneous network of cameras,'' \emph{Journal
  of Mathematical Imaging and Vision}, vol.~41, no. 1-2, pp. 39--58, 2011.

\bibitem{scoop}
\BIBentryALTinterwordspacing
M.~Golbabaee, A.~Alahi, and P.~Vandergheynst, ``Scoop: A real-time sparsity
  driven people localization algorithm,'' \emph{Journal of Mathematical Imaging
  and Vision}, vol.~48, no.~1, pp. 160--175, 2014. [Online]. Available:
  \url{http://dx.doi.org/10.1007/s10851-012-0405-4}
\BIBentrySTDinterwordspacing

\bibitem{peng:mbn}
\BIBentryALTinterwordspacing
P.~Peng, Y.~Tian, Y.~Wang, J.~Li, and T.~Huang, ``Robust multiple cameras
  pedestrian detection with multi-view bayesian network,'' \emph{Pattern
  Recogn.}, vol.~48, no.~5, pp. 1760--1772, May 2015. [Online]. Available:
  \url{http://dx.doi.org/10.1016/j.patcog.2014.12.004}
\BIBentrySTDinterwordspacing

\bibitem{Ge2010}
W.~Ge and R.~T. Collins, ``Crowd detection with a multiview sampler,'' in
  \emph{European Conference on Computer Vision}, 2010.

\bibitem{inception2015}
C.~Szegedy, W.~Liu, Y.~Jia, P.~Sermanet, S.~Reed, D.~Anguelov, D.~Erhan,
  V.~Vanhoucke, and A.~Rabinovich, ``Going deeper with convolutions,'' in
  \emph{Conference on Computer Vision and Pattern Recognition}, 2015.

\bibitem{alexnet2012}
A.~Krizhevsky, I.~Sutskever, and G.~Hinton, ``Imagenet classification with deep
  convolutional neural networks,'' in \emph{Neural Information Processing
  Systems (NIPS)}, 2012.

\bibitem{imagenet_cvpr09}
J.~Deng, W.~Dong, R.~Socher, L.-J. Li, K.~Li, and L.~Fei-Fei, ``{ImageNet: A
  Large-Scale Hierarchical Image Database},'' in \emph{CVPR09}, 2009.

\bibitem{PMT}
P.~Doll\'ar, ``{P}iotr's {C}omputer {V}ision {M}atlab {T}oolbox ({PMT}),''
  \url{https://github.com/pdollar/toolbox}.

\bibitem{Benenson2015}
R.~Benenson, M.~Omran, J.~Hosang, and B.~Schiele, ``Ten years of pedestrian
  detection, what have we learned?'' in \emph{European Conference on Computer
  Vision}, 2015.

\bibitem{pets}
J.~Ferryman and A.~Shahrokni, ``Pets2009: Dataset and challenge,'' in
  \emph{2009 Twelfth IEEE International Workshop on Performance Evaluation of
  Tracking and Surveillance}, Dec 2009, pp. 1--6.

\bibitem{RLCdataset}
``{EPFL-RLC} data-set,'' \url{http://cvlab.epfl.ch/data/rlc}.

\bibitem{metric}
R.~Kasturi, D.~Goldgof, P.~Soundararajan, V.~Manohar, J.~Garofolo, R.~Bowers,
  M.~Boonstra, V.~Korzhova, and J.~Zhang, ``Framework for performance
  evaluation of face, text, and vehicle detection and tracking in video: Data,
  metrics, and protocol,'' \emph{IEEE Transactions on Pattern Analysis and
  Machine Intelligence}, vol.~31, no.~2, pp. 319--336, Feb 2009.

\bibitem{torch}
R.~Collobert, K.~Kavukcuoglu, and C.~Farabet, ``Torch7: A matlab-like
  environment for machine learning,'' in \emph{BigLearn, NIPS Workshop}, 2011.

\bibitem{adam}
\BIBentryALTinterwordspacing
D.~P. Kingma and J.~Ba, ``Adam: {A} method for stochastic optimization,''
  \emph{CoRR}, vol. abs/1412.6980, 2014. [Online]. Available:
  \url{http://arxiv.org/abs/1412.6980}
\BIBentrySTDinterwordspacing

\bibitem{adadelta}
\BIBentryALTinterwordspacing
M.~D. Zeiler, ``{ADADELTA:} an adaptive learning rate method,'' \emph{CoRR},
  vol. abs/1212.5701, 2012. [Online]. Available:
  \url{http://arxiv.org/abs/1212.5701}
\BIBentrySTDinterwordspacing

\bibitem{rmsprop}
\BIBentryALTinterwordspacing
T.~Tieleman and G.~Hinton, ``{RMSprop Gradient Optimization}.'' [Online].
  Available:
  \url{http://www.cs.toronto.edu/\~tijmen/csc321/slides/lecture\_slides\_lec6.pdf}
\BIBentrySTDinterwordspacing

\end{thebibliography}

\checknbdrafts
\end{document}